%% file: main.tex
\documentclass[10pt,twocolumn,letterpaper]{article}

\usepackage{wacv}              

\usepackage{graphicx}
\usepackage{amsmath}
\usepackage{amssymb}
\usepackage{booktabs}

\usepackage{bbm}
\usepackage{animate}
\usepackage{array}
\usepackage{caption}
\usepackage{mathtools}
\usepackage[accsupp]{axessibility} 

\usepackage[pagebackref,breaklinks,colorlinks]{hyperref}

\usepackage[capitalize]{cleveref}
\crefname{section}{Sec.}{Secs.}
\Crefname{section}{Section}{Sections}
\Crefname{table}{Table}{Tables}
\crefname{table}{Tab.}{Tabs.}

\def\wacvPaperID{74} 
\def\confName{WACV}
\def\confYear{2024}

\begin{document}

\title{FPGAN-Control: A Controllable Fingerprint Generator for Training with Synthetic Data}

\author{\textbf{Alon Shoshan}$^1$ \qquad \textbf{Nadav Bhonker}$^1$ \qquad
\textbf{Emanuel Ben Baruch}$^1$ \qquad
\textbf{Ori Nizan}$^2$ \\
\textbf{Igor Kviatkovsky}$^1$ \qquad \textbf{Joshua Engelsma}$^1$ \qquad \textbf{Manoj Aggarwal}$^1$ \qquad \textbf{G\'{e}rard Medioni}$^1$ \\
$^1$Amazon \qquad
$^2$Technion - Israel Institute of Technology \qquad
}

\maketitle

\input{abstract.tex}

\input{introduction.tex}

\input{related_work.tex}

\input{proposed_approach.tex}

\input{experiments.tex}

\input{conclusions.tex}

\newpage

{\small
\bibliographystyle{ieee_fullname}
\bibliography{egbib}
}

\end{document}

%% file: abstract.tex
\begin{abstract}
Training fingerprint recognition models using synthetic data has recently gained increased attention in the biometric community as it alleviates the dependency on sensitive personal data.
Existing approaches for fingerprint generation are limited in their ability to generate diverse impressions of the same finger, a key property for providing effective data for training recognition models.
To address this gap, we present FPGAN-Control, an identity preserving image generation framework which enables control over the fingerprint's image appearance (e.g., fingerprint type, acquisition device, pressure level) of generated fingerprints.
We introduce a novel appearance loss that encourages disentanglement between the fingerprint's identity and appearance properties.
In our experiments, we used the publicly available NIST SD302 (N2N) dataset for training the FPGAN-Control model.
We demonstrate the merits of FPGAN-Control, both quantitatively and qualitatively, in terms of identity preservation level, degree of appearance control, and low synthetic-to-real domain gap.
Finally, training recognition models using only synthetic datasets generated by FPGAN-Control lead to recognition accuracies that are on par or even surpass models trained using real data.
To the best of our knowledge, this is the first work to demonstrate this.
\end{abstract}

%% file: introduction.tex
\section{Introduction}
\label{sec:introduction}
\input{figures/teaser}
Within the past few years the biometric community has shown an increased interest in the use of synthetic data for recognition system development~\cite{engelsma2022printsgan, boutros2022sface, bae2023digiface}. 
This is due to two primary reasons.
First, the state-of-the-art tools for photo-realistic image generation have seen a significant leap in terms of image quality~\cite{karras2020analyzing, karras2021alias, dhariwal2021diffusion,rombach2022high,sauer2022stylegan} and the level of control over the generated output~\cite{shoshan2021gan, deng2020disentangled,KowalskiECCV2020,Tewari_2020_CVPR}. 
While the former reduces the synthetic-to-real domain gap, the latter ensures biometric identity uniqueness and preservation under controllable intra-class variations, enabling the use of synthesized data for both training and evaluation.
Second, recent privacy and ethical concerns regarding the use of existing datasets~\cite{fontanillo2022synthetic} have encouraged researchers to consider replacing real biometric data with synthetic data.

While several efforts have seen success in training models using synthetic data in the face recognition domain~\cite{qiu2021synface,boutros2022sface,bae2023digiface}, the usage of synthetic data for training recognition models in the fingerprints domain has started to gain attention only recently.
One reason for this might be, that while several fingerprint generators are available~\cite{mistry2020fingerprint, bahmani2021high, cao2018fingerprint, wyzykowski2021level}, they lack the ability to generate different impressions for a newly generated identity. 
To tackle this issue, PrintsGAN~\cite{engelsma2022printsgan}, a method for generating novel identities along with their identity preserving variations has been proposed recently.
Although very useful, PrintsGAN relies on a complex three-step generation process focusing on a specific type of fingerprint, which limits the appearance variability in a general sense. 
In addition, the approach lacks the ability to control the generated fingerprint appearance attributed to: fingerprint type (rolled or slap), scanner type, moisture and pressure levels \etc.
Finally, the PrintsGAN generator is not publicly available and only a sample set of 35,000 identities was released.

In this work, we propose an end-to-end GAN based learning scheme for the task of fingerprint image generation, which we name FPGAN-Control (\textbf{F}inger\textbf{P}rint \textbf{GAN}-Control).
In particular, our approach enables control over the appearance of the generated fingerprint images while preserving the biometric identity information.
We rely on the GAN-control framework~\cite{shoshan2021gan} developed initially for controllable and identity-preserving face image generation.

While it is intuitive to define controllable facial characteristics (\eg, expression, age, hair style) and acquisition properties (\eg, orientation, illumination), it is less straightforward for the domain of biometric fingerprints. 
Thus, to address possible variations in impressions of the same finger, we define a single generic ``appearance'' property which encompasses many of the possible impression variations.
To this end, we propose a novel and interpretable appearance loss to enforce disentanglement between the fingerprint identity and its appearance in the GAN's latent space. 
In particular, we apply a smoothing kernel and downsample the generated images in each training batch to filter out their high-frequencies while retaining the appearance properties. 
Then, we encourage similarity between blurred images generated using an identical appearance latent, while separating those generated by different appearance latents.
This way, we enable control over the appearance of the generated fingerprints.
We visualize the control capabilities of the fingerprint appearance while preserving its identity in Figure~\ref{fig:teaser}.

Finally, we use fingerprints generated with FPGAN-Control to train recognition models.
We empirically establish that training fingerprint recognition models using only synthetic identities results in accuracy levels that are comparable and even surpassing those of models trained with real data.
To the best of our knowledge, our method is the first to achieve this in the fingerprints domain.
This allows to avoid relying on sensitive personal data, addressing common privacy and security concerns, highly valued by the biometric community.
To facilitate further research, we will release our code and pretrained models. 


To summarize, our contributions include:
\begin{enumerate}
    \item We introduce FPGAN-Control, an end-to-end training method for controllable fingerprint image generation.
    FPGAN-Control is designed with disentangled latent space in mind to allow generation of novel fingerprint identities along with a variety of impressions.
    \item We propose an intuitive dedicated appearance loss, operating on the fingerprint image's low frequencies. 
    This loss is crucial for disentangling the generator's latent space.
    \item We train recognition models using purely synthetic data, generated by FPGAN-Control, reaching or surpassing the performance of models trained on real data.
    \item Our code and models will be publicly released to facilitate privacy preserving research in the domain of fingerprint recognition. 
\end{enumerate}

%% file: figures/teaser.tex
\begin{figure}
\centering
\animategraphics[loop,autoplay,width=0.85\linewidth]{10}{gifs/gif2/frame-}{0}{100}
\caption{\textbf{Traversing the appearance space of FPGAN-Control.}
We present an animation of four fingerprints generated by FPGAN-Control with the following properties:
(a) each of four animated fingerprints belongs to a unique synthetic identity which is preserved throughout the animation;
(b) at every moment the appearance of each fingerprint is shared; and
(c) the shared appearance gradually changes over time.
[Animated figure, please view in Acrobat Reader].}
\label{fig:teaser}
\end{figure}








%% file: related_work.tex
\input{figures/fig_app_loss}
\section{Related Work}
\label{sec:related-work}
The scarcity of publicly available fingerprint datasets has led to increased interest in developing methods to synthesize fingerprints.
Earlier methods relied heavily on ``hand-crafted'' solutions based on the available knowledge on fingerprints, \eg, they leverage the studied behavior of the friction ridge patterns comprised of interwoven ridges, valleys, minutiae points, and pores~\cite{cappelli2002synthetic, zhao2012fingerprint, johnson2013texture, ansari2011generation}.
In more recent years, as Generative Adversarial Networks (GANs)~\cite{GoodfellowPMXWOCB14} have proliferated into a host of photo-realistic image synthesis algorithms, researchers have turned to GANs or ``learning based methods''~\cite{minaee2018finger,bontrager2018deepmasterprints,attia2019fingerprint,riazi2020synfi,fahim2020lightweight,bahmani2021high,sams2022hq}, to generate much more realistic fingerprints than the older hand-crafted approaches. 
Although most of these methods are able to produce high-quality fingerprint images, they suffer from two main deficiencies.
First, the methods are only able to generate a single, unique image for each synthetic fingerprint identity.
Hence, they are not able to model the intra-class variability.
Second, many of these synthesize only patches of fingerprints, rather than full-fingerprints.
These limitations motivated methods which introduce control over the generated fingerprints, allowing the generation of multiple impressions per fingerprint identity~\cite{
wyzykowski2021level,
engelsma2022printsgan, grosz2022spoofgan,priesnitz2022syncolfinger,wyzykowski2023synthetic}. 

These methods usually consist of multiple stages, \eg, binary fingerprint generation, distortion, and a GAN to render the binary fingerprint to a realistic fingerprint impression.
The need to apply many stages, adds significant complexity to the systems and risk when adapting to new datasets and domains.
Most importantly, while all previous work agree that there's a need for synthetic data to conduct research, only two~\cite{engelsma2022printsgan,grosz2022spoofgan} of the above methods evaluate the performance of models trained using their synthetic data.
Since Grosz~\etal~\cite{grosz2022spoofgan} proposes a generator of spoof images, the method closest to ours is PrintsGAN~\cite{engelsma2022printsgan}.

Our method consists of a single training stage based on GAN-Control~\cite{shoshan2021gan} with a disentangled latent space. 
We incorporate a novel appearance loss to enable the control of both identity and appearance details.
In addition, we do not rely on any hand-crafted image synthesizer or other proprietary algorithms.
Our method benefits from the advantages of previous learning-based approaches while not suffering from their deficiencies. 
In particular, our approach is able to generate multiple impressions of a single finger while preserving its identity; 
our method consists of a single network trained end-to-end;
and training recognition models using synthetic data generated by FPGAN-Control reaches performance comparable to that of models trained on real data.

Worth noting are the recent efforts made towards training face recognition models using synthetic data ~\cite{qiu2021synface,boutros2022sface,bae2023digiface,kortylewski2018training,kortylewski2019analyzing,luo2021fa,yang2022heterogeneous}.
For instance, in~\cite{qiu2021synface}, DiscoFaceGAN~\cite{deng2020disentangled} was used to generate synthetic face identities for training, and proposed to deploy identity mixup and domain mixup to mitigate the domain gap between real and synthetic images.
In~\cite{bae2023digiface} the authors released a large-scale synthetic dataset created by rendering 3D face models, containing 1.22M images of 110K identities and utilized it for training.
While demonstrating appealing accuracy results for face recognition, it is not trivial to adapt them to the fingerprint domain, as they are specifically handcrafted for the face domain, \eg, using 3D face models~\cite{bae2023digiface} or training GANs with domain specific losses~\cite{deng2020disentangled}).







%% file: figures/fig_app_loss.tex
\begin{figure*}
\centering
\begin{subfigure}[t]{0.33\linewidth}
\centering
\includegraphics[height=0.755\linewidth]{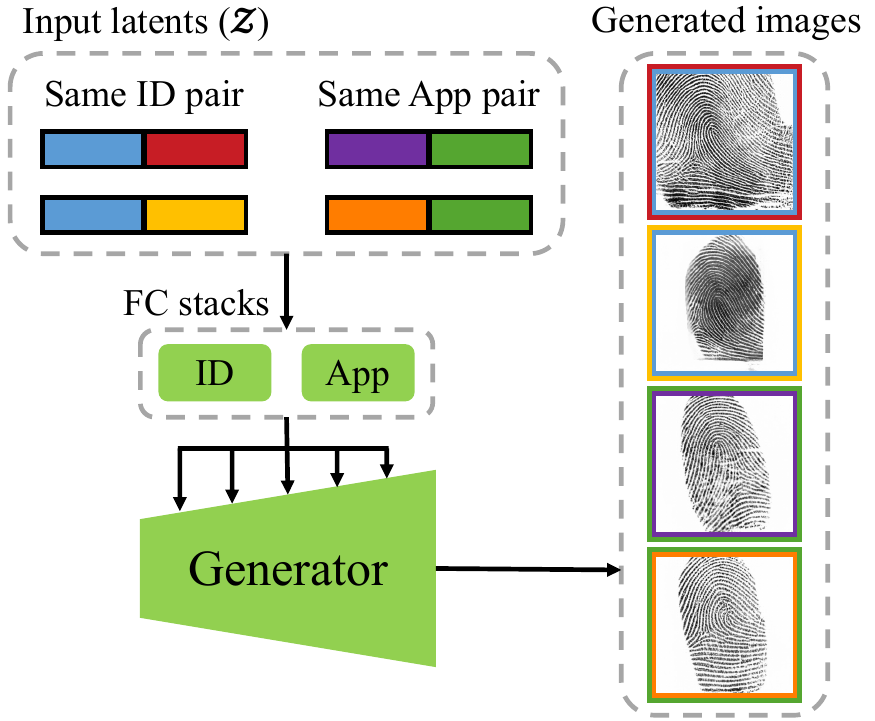}
\caption{Training batch generation.}
\label{fig:training_batch_generation}
\end{subfigure}
\begin{subfigure}[t]{0.33\linewidth}
\centering
\includegraphics[height=0.755\linewidth]{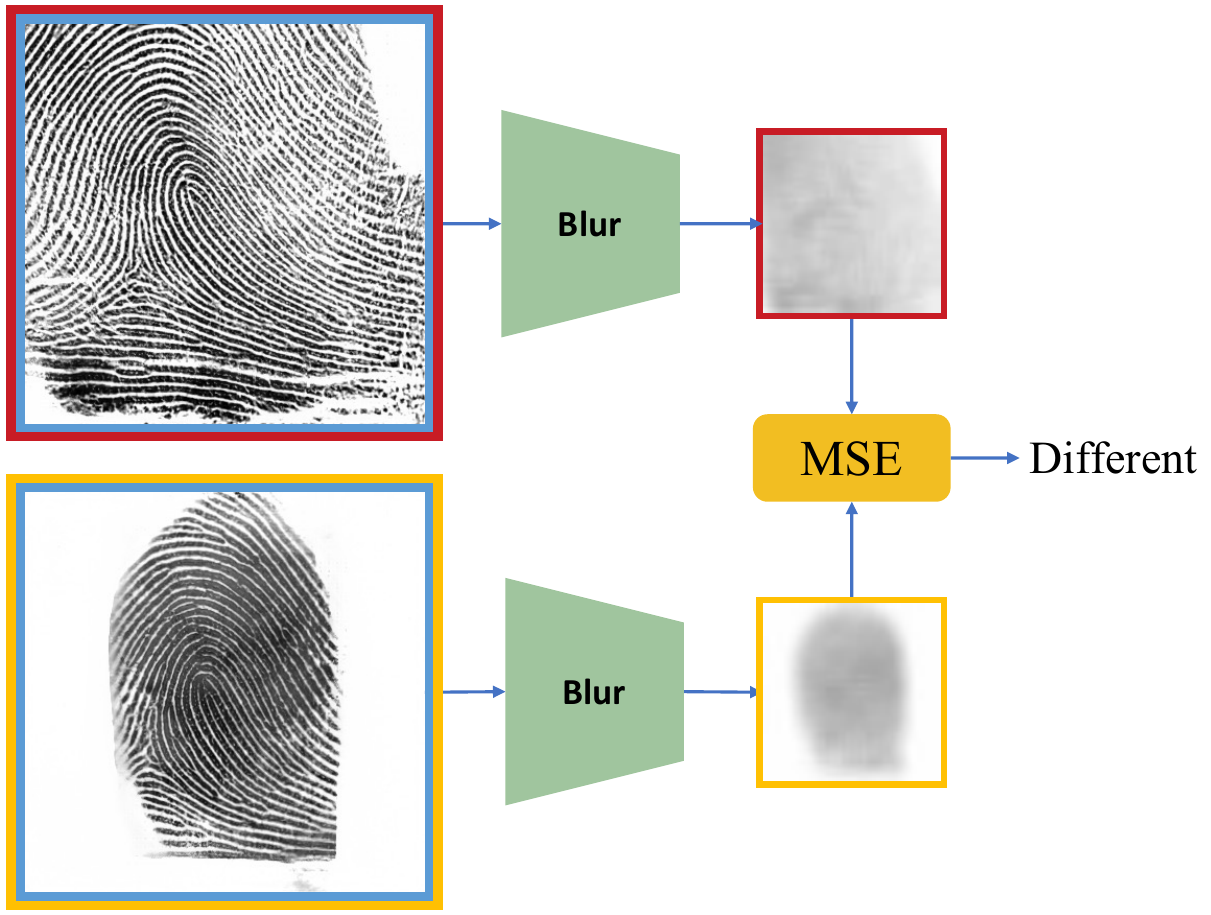}
\caption{Different appearance with same ID.}
\label{fig:different_appearance}
\end{subfigure}
\begin{subfigure}[t]{0.33\linewidth}
\centering
\includegraphics[height=0.755\linewidth]{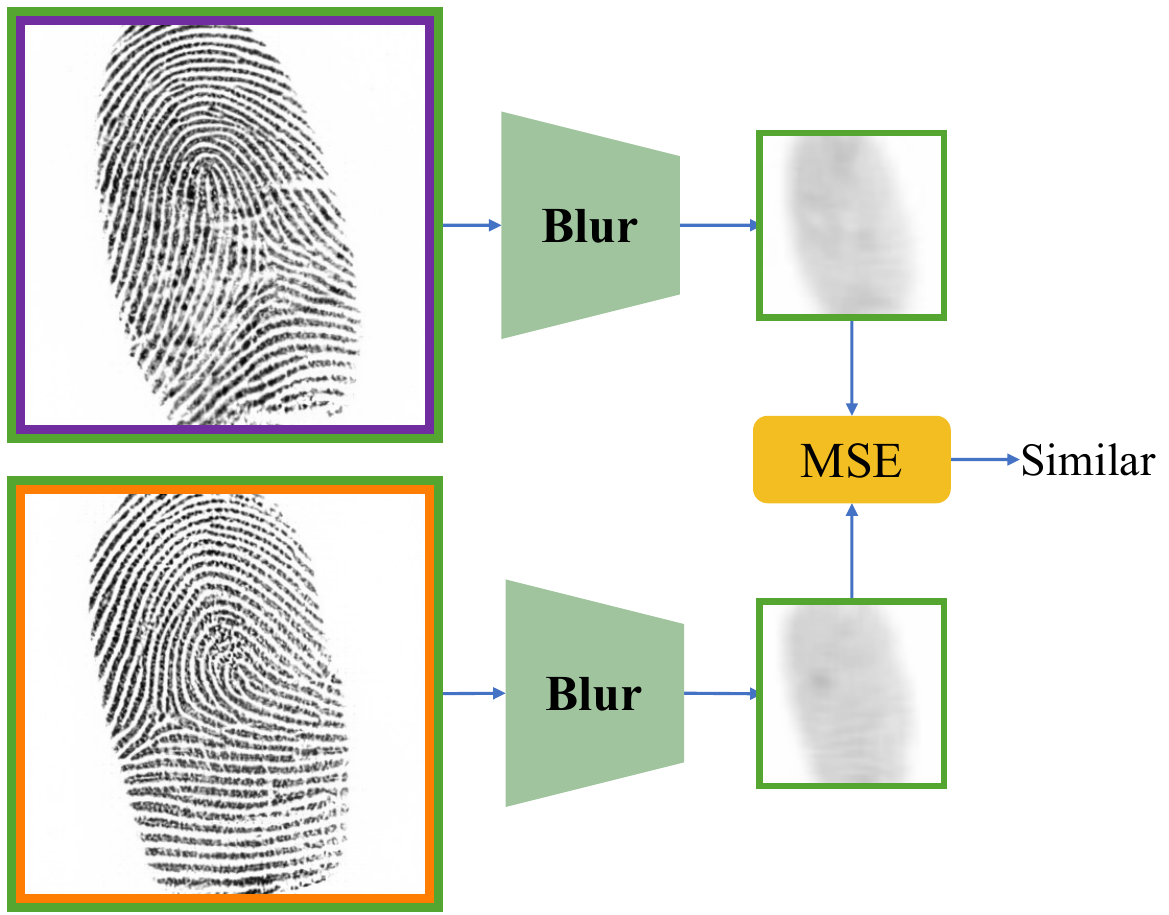}
\caption{Same appearance with different ID.}
\label{fig:same_appearance}
\end{subfigure}

\caption{
\textbf{Overview of the proposed FPGAN-Control.}
In each training batch (a), both same ID pairs and same appearance pairs are generated.
Same ID pairs have the same ID latent vector while same appearance pairs have the same appearance latent vector.
The color of the inner image border corresponds to the fingerprint ID and the color of the outer border corresponds to the fingerprint appearance. 
Each image in the batch is blurred and downsampled, effectively removing it's barometric features while still obtaining many of its appearance features.
Blurred images with different appearance latents are pushed one from another (b), while blurred images with the same appearance latent are pulled towards each other (c).
}
\vspace{-1.0em}
\label{fig:app_loss}
\end{figure*}

%% file: proposed_approach.tex
\section{Proposed Approach}
\label{sec:proposed-approach}
In this section we present our framework for training FPGAN-Control, a fingerprint image generation model that allows generating novel biometric identities (IDs) and controlling their appearance variations while preserving the ID.
We define the appearance as the image properties that are not related to the fingerprint ID, such as fingerprint type, acquisition device, moisture and pressure levels, \etc.

\subsection{Identity-appearance disentangling}
In the design of our approach we rely on the core blocks of GAN-Control~\cite{shoshan2021gan}.
GAN-Control proposes a general two-phase solution for training explicit controllable GANs.
In the first phase, we train a disentangled GAN using a set of contrastive losses.
As a result, the latent space of the trained GAN is divided into subspaces, each encoding a different image property.
The second phase is responsible for enabling explicit control over the generated image by adding property-specific encoders to each subspace of the GAN.
While an explicit control over image attributes is useful for content creation, such fine-grained control is not required for the purpose of data generation for recognition model training.  
Thus, to train FPGAN-Control we adopt only the first phase, resulting in a disentanglement between identity and appearance.

We define the latent space, $\mathcal{Z}$, of FPGAN-Control as a combination of two subspaces $\mathcal{Z}^{id}$ and $\mathcal{Z}^{app}$, associated with the generated fingerprint ID and appearance, respectively ($\mathcal{Z}=\mathcal{Z}^{id} \times \mathcal{Z}^{app}$).  
Thus, a latent $\mathbf{z} \in \mathcal{Z}$ is the concatenation of the sub-vectors $\mathbf{z}^{id}$ and $\mathbf{z}^{app}$, each of dimension 256.
Instead of using a single 8-layered MLP, as done in StyleGAN2~\cite{karras2020analyzing}, we allocate a separate 8-layered MLP for each subspace. 
Each training batch contains pairs of same-ID latents (latent vectors with equal identity, $\mathbf{z}^{id}$, and different appearance, $\mathbf{z}^{app}$) and same-appearance latents (equal appearance, $\mathbf{z}^{app}$ and different ID, $\mathbf{z}^{id}$), see Figure~\ref{fig:training_batch_generation}.
In addition to the StyleGAN2's original adversarial loss, all image pairs are penalized by a weighted combination of contrastive ID and appearance losses ($l_{id}$ and $l_{app}$):
\begin{equation}
L_c = \sum_{\substack{\mathbf{z}_i,\mathbf{z}_j \in B\\ i \ne j}} l_{id}(\mathbf{z}_i,\mathbf{z}_j) + w_{app} \cdot l_{app}(\mathbf{z}_i,\mathbf{z}_j),     
\end{equation}
where $B = \{ \mathbf{z}_i \}_{i=1}^{N_B} $ denotes all latent vectors in the training batch of size $N_B$.
Each of the loss components, $l_{id}$ and $l_{app}$, has a generic form depending on the corresponding distance function, $d_{id}$ and $d_{app}$:
\begin{equation}
\small
l_k(\mathbf{z}_i,\mathbf{z}_j)\!=\! 
\begin{cases}
    \frac{1}{C_k^{+}} \max{(d_k(\mathbf{\cal I}_i,\mathbf{\cal I}_j) - \tau_k^+, 0)},&  \mathbf{z}_i^k\!=\!\mathbf{z}_j^k \\
    \frac{1}{C_k^\texttt{-}} \max{(\tau_k^- - d_k(\mathbf{\cal I}_i,\mathbf{\cal I}_j), 0)},& \text{otherwise}
\end{cases}
\end{equation}
where $k\in\{id, app\}$, $\tau_k^{\pm}$ are ID or appearance thresholds associated with same and different sub-vectors and $C_k^{\pm}$ are normalizing constants. 
The key part of this scheme is in the selection of appropriate distance functions, $d_{id}$ and $d_{app}$. 
We define $d_{id}$ as the cosine distance between two embedding vectors extracted by a pre-trained fingerprint recognition model, $\theta_{id}$.
Unfortunately, there is no available identity invariant metric for computing the appearance distance, $d_{app}$, between two images.
To address this, we developed an appearance distance focusing on the dissimilarity between a pair of images in the low-frequency domain.

\subsection{Appearance distance}
Intuitively, given a fingerprint scan, most of the ID-related biometric features are contained in the image's high frequency components while the appearance-related information is contained in its low frequency components.
We use this observation in the design of our appearance distance function, $d_{app}$. 
Given two fingerprint images $\mathbf{\cal I}_i$ and $\mathbf{\cal I}_j$, we downsample and blur each image using a Gaussian smoothing filter:
\begin{equation}
\Tilde{\mathbf{\cal I}} = \text{resize}(\mathbf{\cal I}) *  h(\sigma, n),  
\end{equation}
where $\text{resize}(\cdot)$ is a bi-linear downsampling operation, $h$ is a Gaussian kernel with variance $\sigma$ and kernel size $n$.
The purpose of the blur filter is to remove as much of the ID-dependent biometric features as possible from the fingerprint, while preserving the most important appearance-related information. To measure the appearance distance between $\mathbf{\cal I}_i$ and $\mathbf{\cal I}_j$ we compute the pixelwise Mean Squared Error (MSE) between their processed versions: 
\begin{equation}
d_{app}(\mathbf{\cal I}_i, \mathbf{\cal I}_j) = \text{MSE}(\Tilde{\mathbf{\cal I}_i}, \Tilde{\mathbf{\cal I}_j}).
\label{eq:distance}
\end{equation}
In each training batch we use $d_{app}$ to penalize images with different appearance latents but having similar appearance (Figure~\ref{fig:different_appearance}) and images sharing the same appearance latent but diverging in appearance (Figure~\ref{fig:same_appearance}).
In Section~\ref{sec:experiments} we demonstrate that this intuitive approach is effective in training identity-preserving fingerprint generators with controllable appearance.

\subsection{Removing the first sub-sampling layer of fingerprint recognition models}
Throughout our experimental work, we observed that training fingerprint recognition models using common off-the-shelf architectures (\ie, ResNet, MobileNet, and EfficientNet) lead to unstable and poor results.
We hypothesized that the initial sub-sampling layer of these networks may eliminate valuable fine-grained fingerprint details that are crucial for achieving high discriminative power. 
Therefore, we removed the initial sub-sampling mechanism from all networks.
Specifically, we omitted the first max-pooling layer from ResNet models, and we reduced the stride size of the initial convolutional layer from 2 to 1 in MobileNet and EfficientNet architectures.
With this modification, we achieved increased training stability and a significant improvement in test accuracy. 
We implemented this architectural change in all of the recognition models that were experimented with in the paper.

Table~\ref{tab:id_app_same_id} presents the results obtained by models trained both with and without the initial sub-sampling layer.
As can be seen, training recognition models without the initial sub-sampling layer improves the test accuracy by a large margin for all the tested architectures.

\input{tables/table_w_vs_wo_pooling_layer}

\input{figures/qulitative_app_change}

%% file: tables/table_w_vs_wo_pooling_layer.tex
\begin{table}[t]
    \centering

    \small
    \begin{tabular}{l | c c c c c c c}
    \toprule
                       & R18    & R34    & R50    & R101   & M050   & M100   & Eff-s \\
    \midrule
    \footnotesize w/    & $82.6$ & $80.3$ & $82.2$ & $83.0$ & $90.8$ & $90.2$ & $90.6$ \\
    \footnotesize w/o  & $90.5$ & $89.7$ & $92.0$ & $93.7$ & $93.7$ & $94.2$ & $93.7$ \\
    \bottomrule
    \end{tabular}

    
    \caption{\textbf{Effect of first sub-sampling layer.} TAR@FAR=0.1\% results for recognition models trained with and without (w/, w/o) the first backbone's sub-sampling layer.
    The following backbones where evaluated: ResNet (R), MobileNetV2 (M) and EfficientNetV2 (Eff).}
    \vspace{-1.0em}
    \label{tab:id_app_same_id}
\end{table}

%% file: figures/qulitative_app_change.tex
\begin{figure*}

\centering
\setlength{\fboxsep}{0pt}
\begin{subfigure}[t]{0.48\linewidth}
\renewcommand{\arraystretch}{0}
\begin{tabular}{@{}c@{}c@{}c@{}c@{}c@{}c@{}}
    \footnotesize
    & \tiny ID 1 & \tiny ID 2 & \tiny ID 3 & \tiny ID 4 & \tiny ID 5 \\ [0.1em]
    \raisebox{0.7\totalheight}{\rotatebox[origin=c]{90}{\tiny Appearance 1}} &
    \fbox{\includegraphics[width=0.19\linewidth]{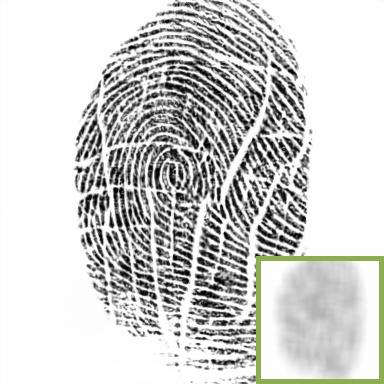}}\hfill &
    \fbox{\includegraphics[width=0.19\linewidth]{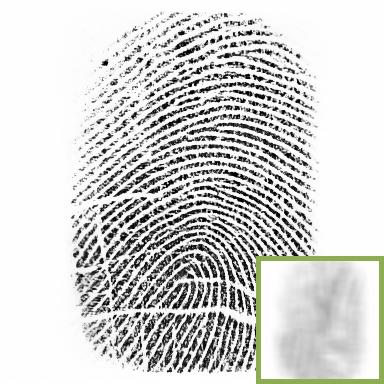}}\hfill &
    \fbox{\includegraphics[width=0.19\linewidth]{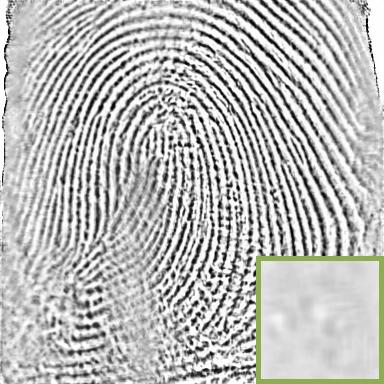}}\hfill &
    \fbox{\includegraphics[width=0.19\linewidth]{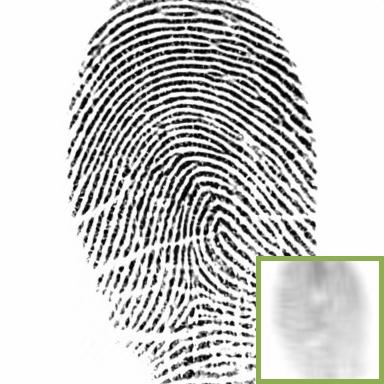}}\hfill &
    \fbox{\includegraphics[width=0.19\linewidth]{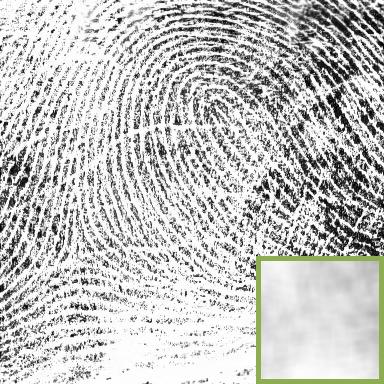}}\hfill \\
    \raisebox{0.7\totalheight}{\rotatebox[origin=c]{90}{\tiny Appearance 2}} &
    \fbox{\includegraphics[width=0.19\linewidth]{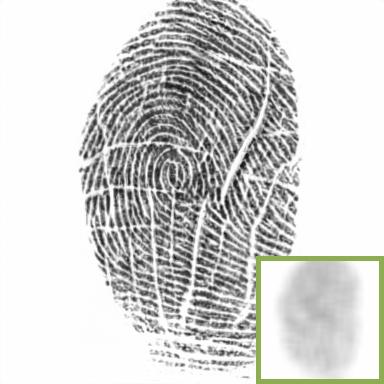}}\hfill &
    \fbox{\includegraphics[width=0.19\linewidth]{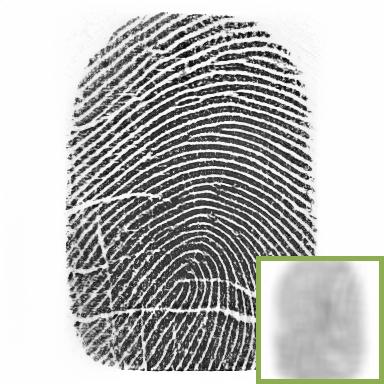}}\hfill &
    \fbox{\includegraphics[width=0.19\linewidth]{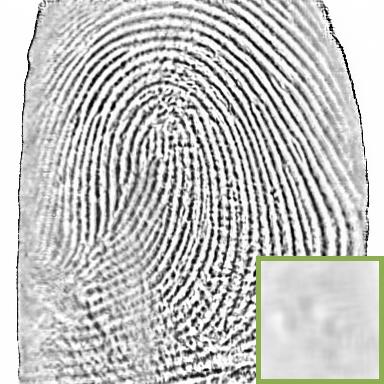}}\hfill &
    \fbox{\includegraphics[width=0.19\linewidth]{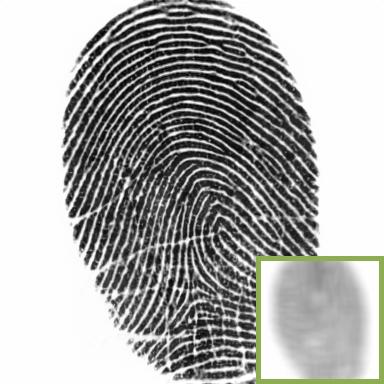}}\hfill &
    \fbox{\includegraphics[width=0.19\linewidth]{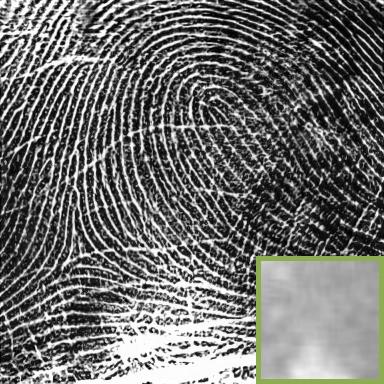}}\hfill \\
    \raisebox{0.7\totalheight}{\rotatebox[origin=c]{90}{\tiny Appearance 3}} &
    \fbox{\includegraphics[width=0.19\linewidth]{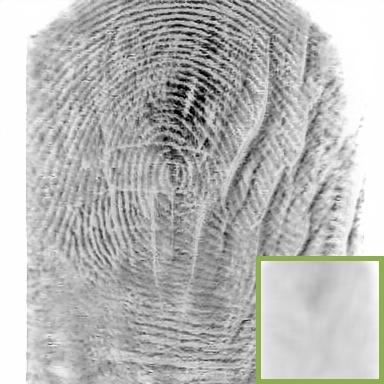}}\hfill &
    \fbox{\includegraphics[width=0.19\linewidth]{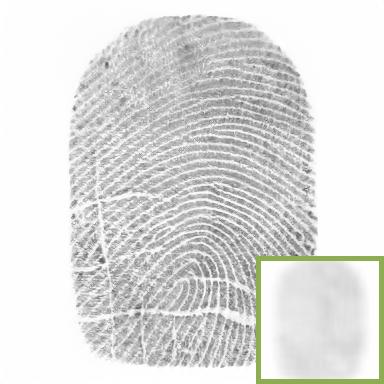}}\hfill &
    \fbox{\includegraphics[width=0.19\linewidth]{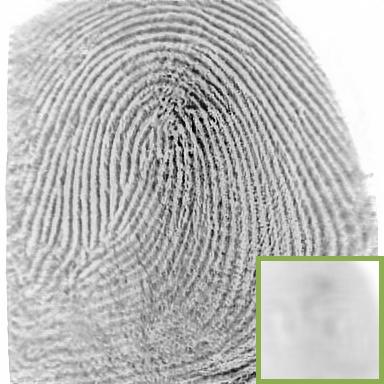}}\hfill &
    \fbox{\includegraphics[width=0.19\linewidth]{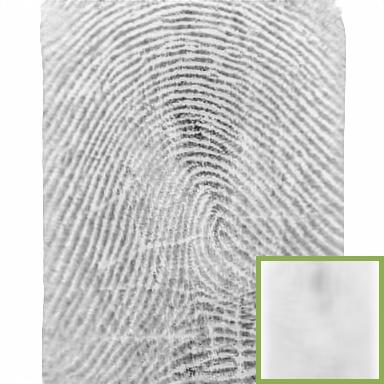}}\hfill &
    \fbox{\includegraphics[width=0.19\linewidth]{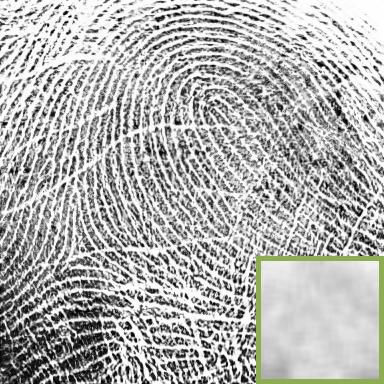}}\hfill \\

    \raisebox{0.7\totalheight}{\rotatebox[origin=c]{90}{\tiny Appearance 4}} &
    \fbox{\includegraphics[width=0.19\linewidth]{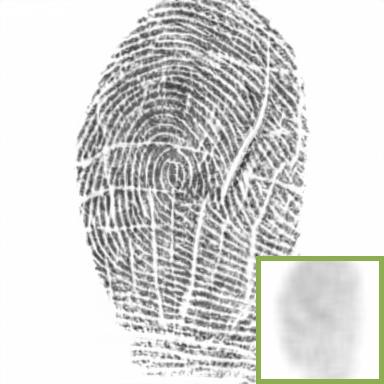}}\hfill &
    \fbox{\includegraphics[width=0.19\linewidth]{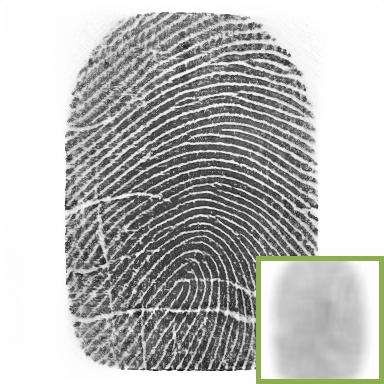}}\hfill &
    \fbox{\includegraphics[width=0.19\linewidth]{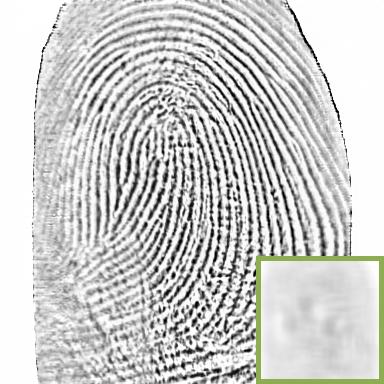}}\hfill &
    \fbox{\includegraphics[width=0.19\linewidth]{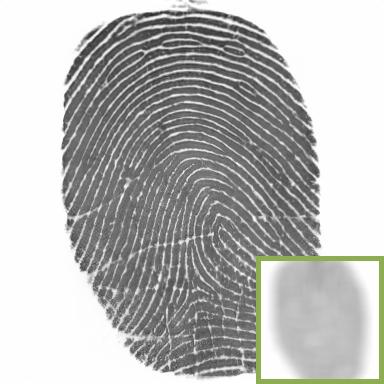}}\hfill &
    \fbox{\includegraphics[width=0.19\linewidth]{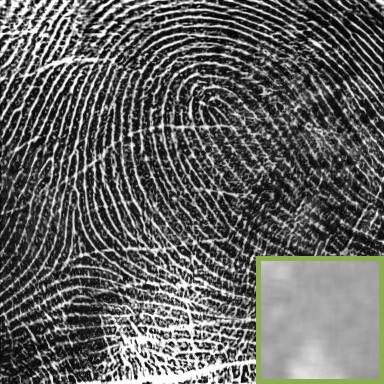}}\hfill \\
\end{tabular}
\caption{$w_{app}=0$.}
\label{fig:qualitative_results_0}
\end{subfigure}
\begin{subfigure}[t]{0.48\linewidth}
\renewcommand{\arraystretch}{0}
\begin{tabular}{@{}c@{}c@{}c@{}c@{}c@{}c@{}}
    \footnotesize
    & \tiny ID 1 & \tiny ID 2 & \tiny ID 3 & \tiny ID 4 & \tiny ID 5 \\ [0.1em]
    \raisebox{0.7\totalheight}{\rotatebox[origin=c]{90}{\tiny Appearance 1}} &
    \fbox{\includegraphics[width=0.19\linewidth]{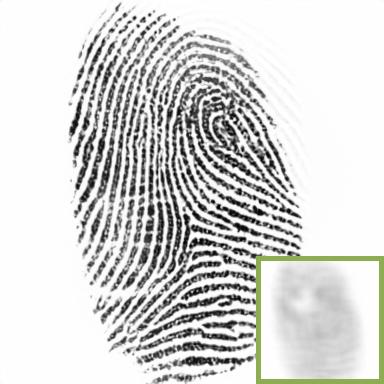}}\hfill &
    \fbox{\includegraphics[width=0.19\linewidth]{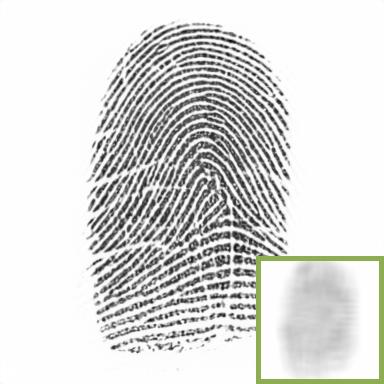}}\hfill &
    \fbox{\includegraphics[width=0.19\linewidth]{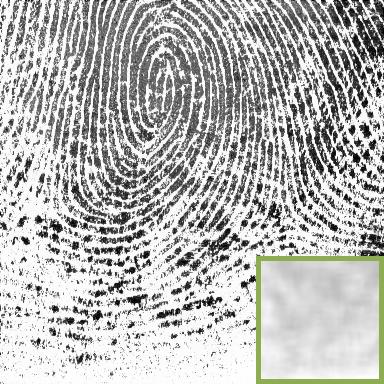}}\hfill &
    \fbox{\includegraphics[width=0.19\linewidth]{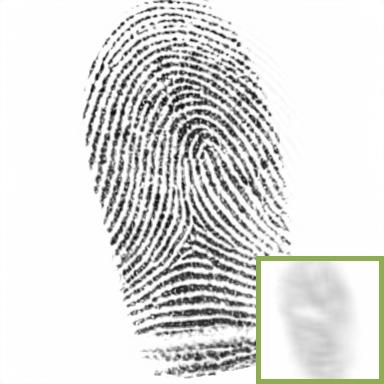}}\hfill &
    \fbox{\includegraphics[width=0.19\linewidth]{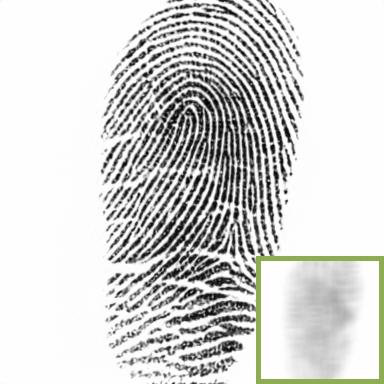}}\hfill \\
    \raisebox{0.7\totalheight}{\rotatebox[origin=c]{90}{\tiny Appearance 2}} &
    \fbox{\includegraphics[width=0.19\linewidth]{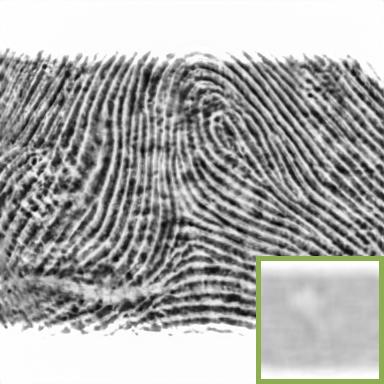}}\hfill &
    \fbox{\includegraphics[width=0.19\linewidth]{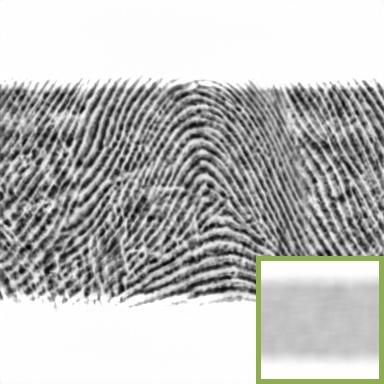}}\hfill &
    \fbox{\includegraphics[width=0.19\linewidth]{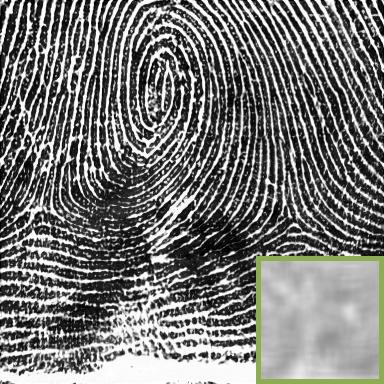}}\hfill &
    \fbox{\includegraphics[width=0.19\linewidth]{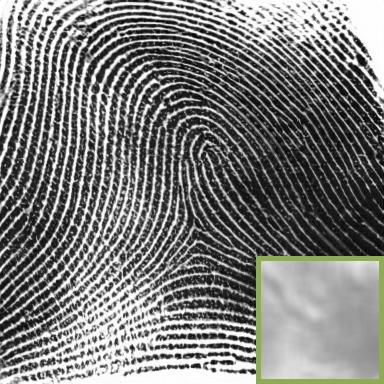}}\hfill &
    \fbox{\includegraphics[width=0.19\linewidth]{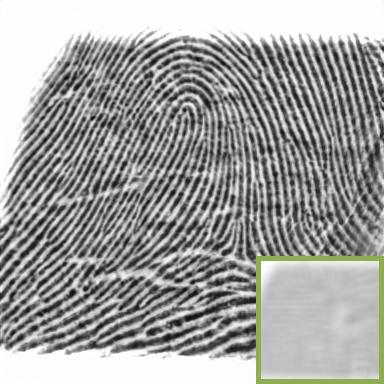}}\hfill \\
    \raisebox{0.7\totalheight}{\rotatebox[origin=c]{90}{\tiny Appearance 3}} &
    \fbox{\includegraphics[width=0.19\linewidth]{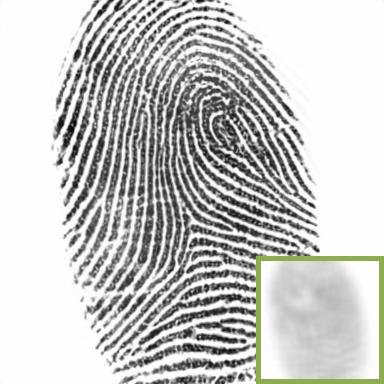}}\hfill &
    \fbox{\includegraphics[width=0.19\linewidth]{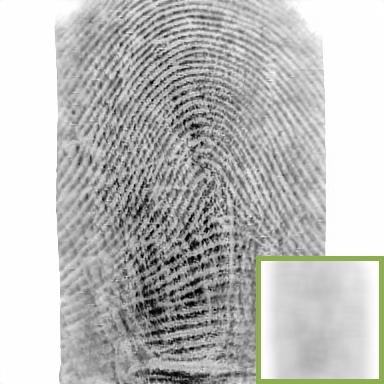}}\hfill &
    \fbox{\includegraphics[width=0.19\linewidth]{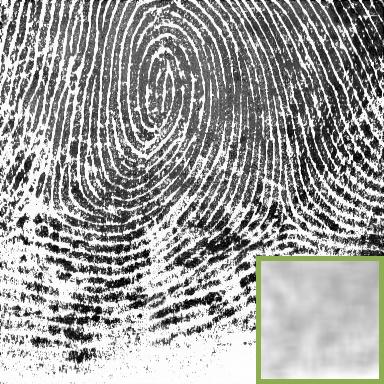}}\hfill &
    \fbox{\includegraphics[width=0.19\linewidth]{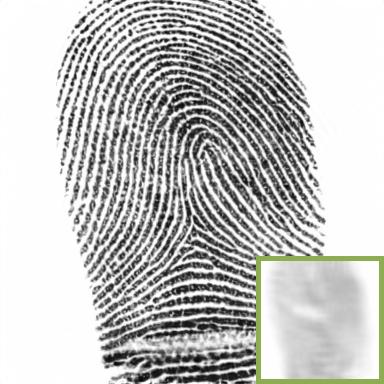}}\hfill &
    \fbox{\includegraphics[width=0.19\linewidth]{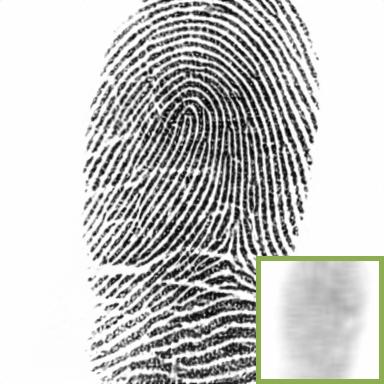}}\hfill \\

    \raisebox{0.7\totalheight}{\rotatebox[origin=c]{90}{\tiny Appearance 4}} &
    \fbox{\includegraphics[width=0.19\linewidth]{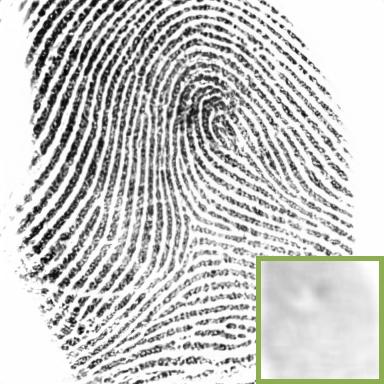}}\hfill &
    \fbox{\includegraphics[width=0.19\linewidth]{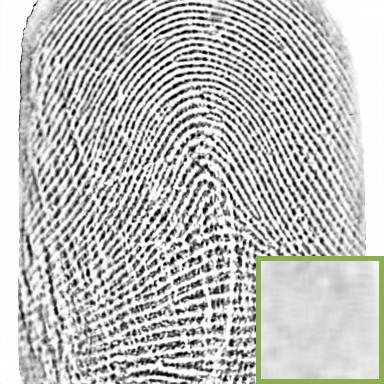}}\hfill &
    \fbox{\includegraphics[width=0.19\linewidth]{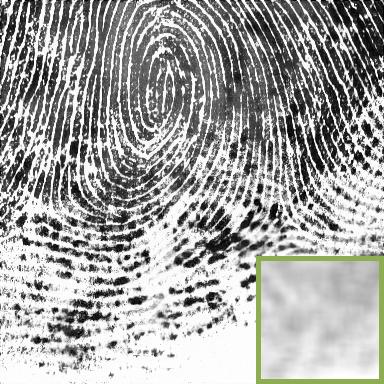}}\hfill &
    \fbox{\includegraphics[width=0.19\linewidth]{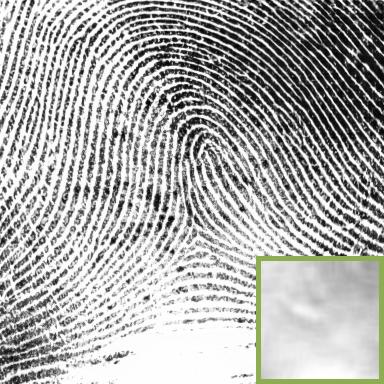}}\hfill &
    \fbox{\includegraphics[width=0.19\linewidth]{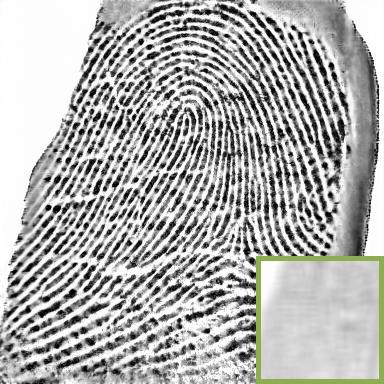}}\hfill \\
\end{tabular}
\caption{$w_{app}=1$.}
\label{fig:qualitative_results_1}
\end{subfigure}

\vspace{10px}

\begin{subfigure}[t]{0.49\linewidth}
\renewcommand{\arraystretch}{0}
\begin{tabular}{@{}c@{}c@{}c@{}c@{}c@{}c@{}}
    \footnotesize
    & \tiny ID 1 & \tiny ID 2 & \tiny ID 3 & \tiny ID 4 & \tiny ID 5 \\ [0.1em]
    \raisebox{0.7\totalheight}{\rotatebox[origin=c]{90}{\tiny Appearance 1}} &
    \fbox{\includegraphics[width=0.19\linewidth]{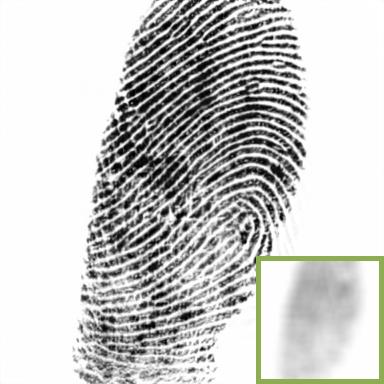}}\hfill &
    \fbox{\includegraphics[width=0.19\linewidth]{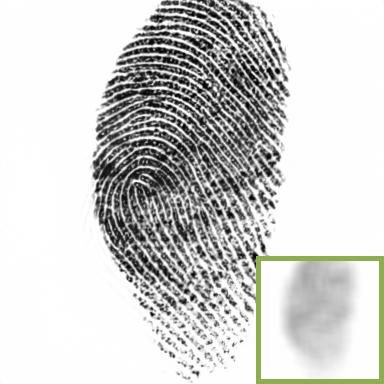}}\hfill &
    \fbox{\includegraphics[width=0.19\linewidth]{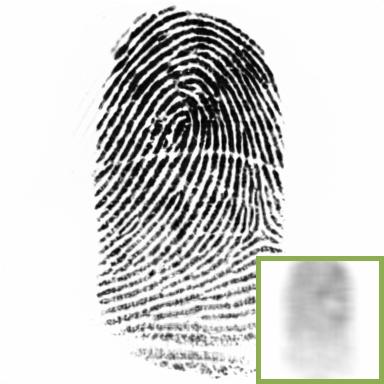}}\hfill &
    \fbox{\includegraphics[width=0.19\linewidth]{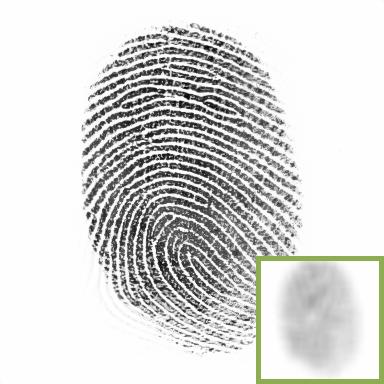}}\hfill &
    \fbox{\includegraphics[width=0.19\linewidth]{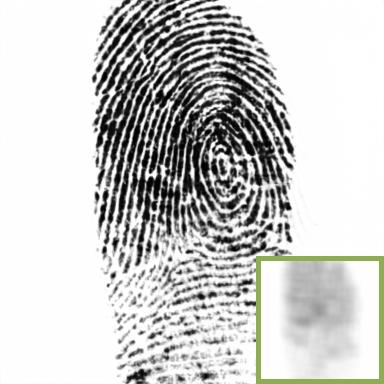}}\hfill \\
    \raisebox{0.7\totalheight}{\rotatebox[origin=c]{90}{\tiny Appearance 2}} &
    \fbox{\includegraphics[width=0.19\linewidth]{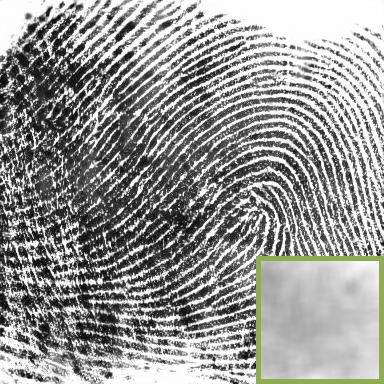}}\hfill &
    \fbox{\includegraphics[width=0.19\linewidth]{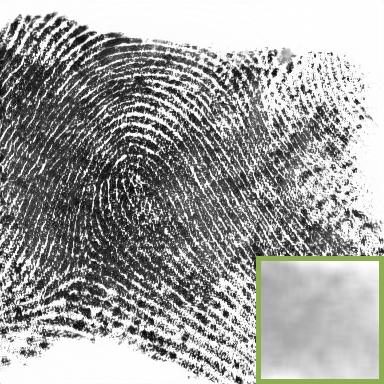}}\hfill &
    \fbox{\includegraphics[width=0.19\linewidth]{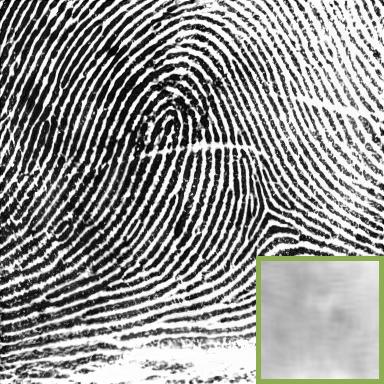}}\hfill &
    \fbox{\includegraphics[width=0.19\linewidth]{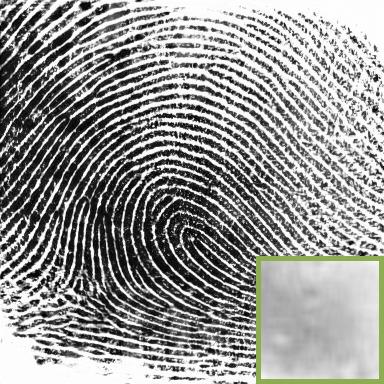}}\hfill &
    \fbox{\includegraphics[width=0.19\linewidth]{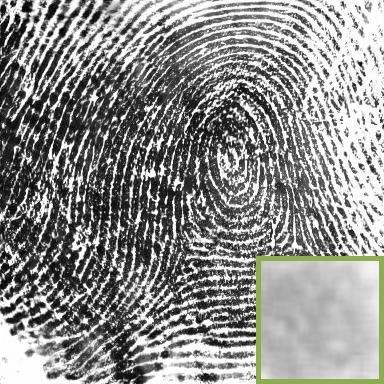}}\hfill \\
    \raisebox{0.7\totalheight}{\rotatebox[origin=c]{90}{\tiny Appearance 3}} &
    \fbox{\includegraphics[width=0.19\linewidth]{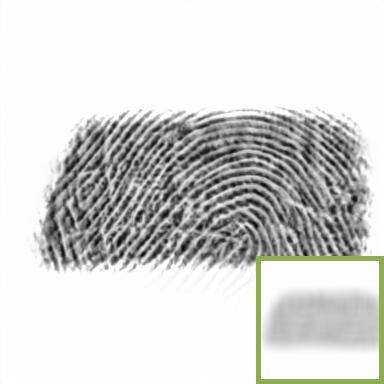}}\hfill &
    \fbox{\includegraphics[width=0.19\linewidth]{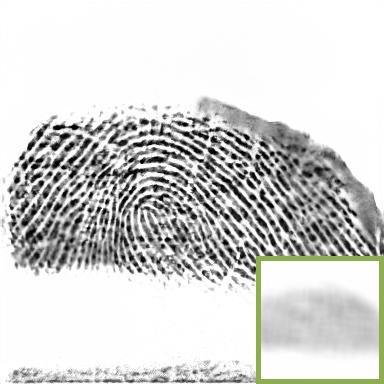}}\hfill &
    \fbox{\includegraphics[width=0.19\linewidth]{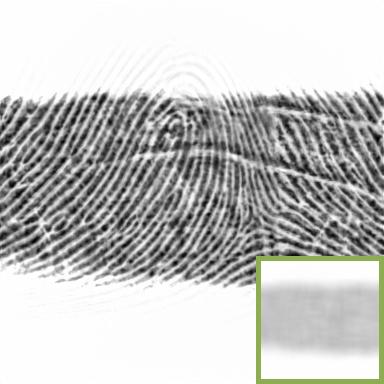}}\hfill &
    \fbox{\includegraphics[width=0.19\linewidth]{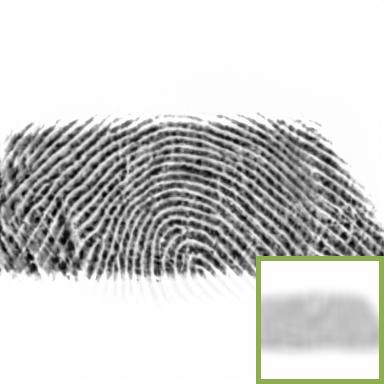}}\hfill &
    \fbox{\includegraphics[width=0.19\linewidth]{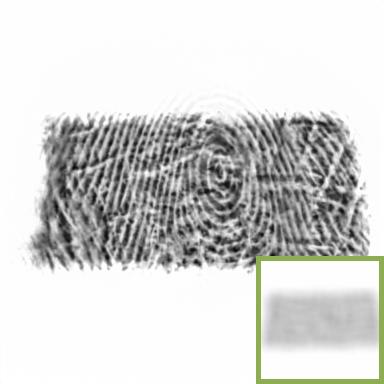}}\hfill \\

    \raisebox{0.7\totalheight}{\rotatebox[origin=c]{90}{\tiny Appearance 4}} &
    \fbox{\includegraphics[width=0.19\linewidth]{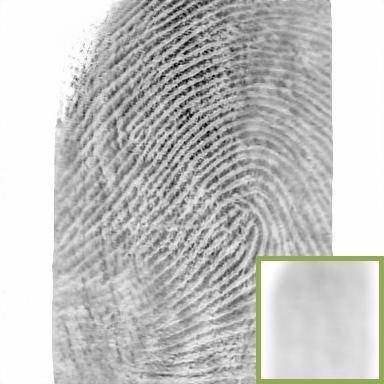}}\hfill &
    \fbox{\includegraphics[width=0.19\linewidth]{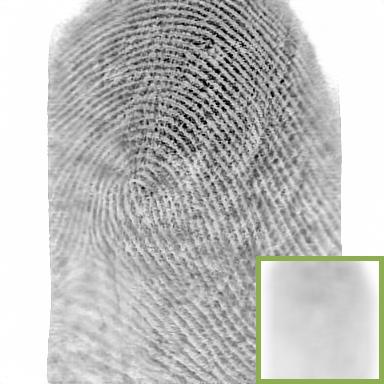}}\hfill &
    \fbox{\includegraphics[width=0.19\linewidth]{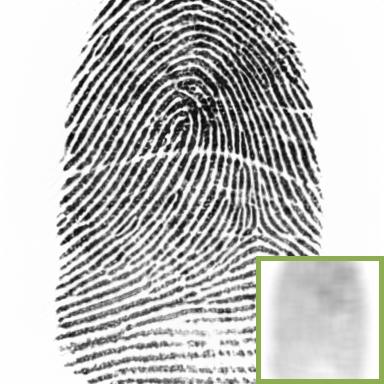}}\hfill &
    \fbox{\includegraphics[width=0.19\linewidth]{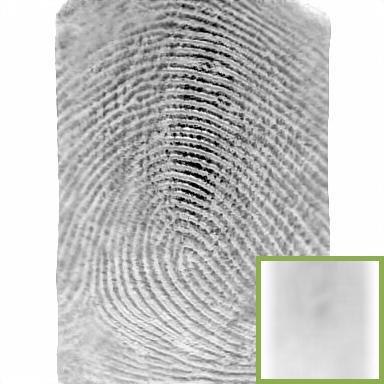}}\hfill &
    \fbox{\includegraphics[width=0.19\linewidth]{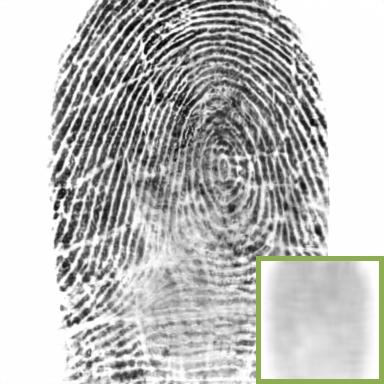}}\hfill \\
\end{tabular}
\caption{$w_{app}=5$.}
\label{fig:qualitative_results_5}
\end{subfigure}
\begin{subfigure}[t]{0.49\linewidth}
\renewcommand{\arraystretch}{0}
\begin{tabular}{@{}c@{}c@{}c@{}c@{}c@{}c@{}}
    \footnotesize
    & \tiny ID 1 & \tiny ID 2 & \tiny ID 3 & \tiny ID 4 & \tiny ID 5 \\ [0.1em]
    \raisebox{0.7\totalheight}{\rotatebox[origin=c]{90}{\tiny Appearance 1}} &
    \fbox{\includegraphics[width=0.19\linewidth]{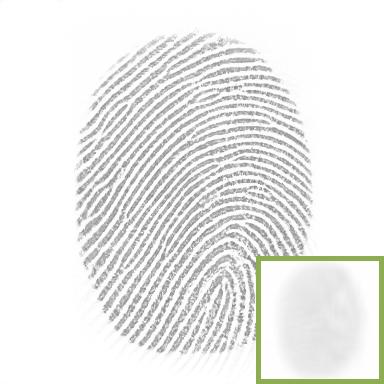}}\hfill &
    \fbox{\includegraphics[width=0.19\linewidth]{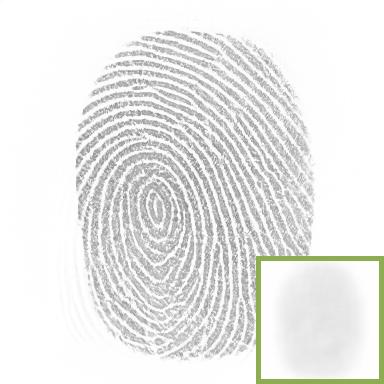}}\hfill &
    \fbox{\includegraphics[width=0.19\linewidth]{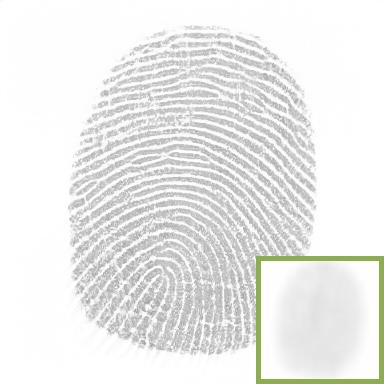}}\hfill &
    \fbox{\includegraphics[width=0.19\linewidth]{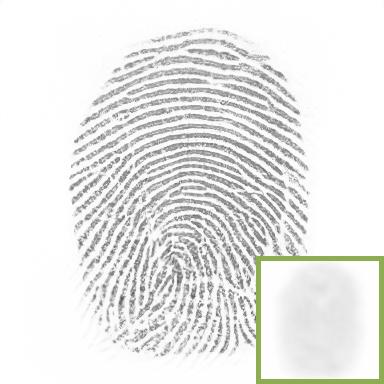}}\hfill &
    \fbox{\includegraphics[width=0.19\linewidth]{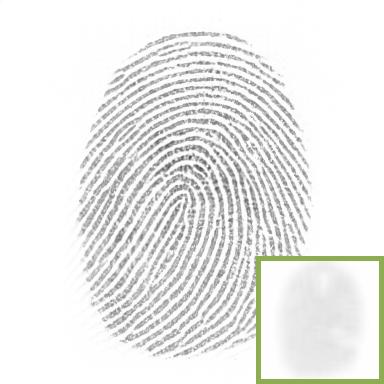}}\hfill \\
    \raisebox{0.7\totalheight}{\rotatebox[origin=c]{90}{\tiny Appearance 2}} &
    \fbox{\includegraphics[width=0.19\linewidth]{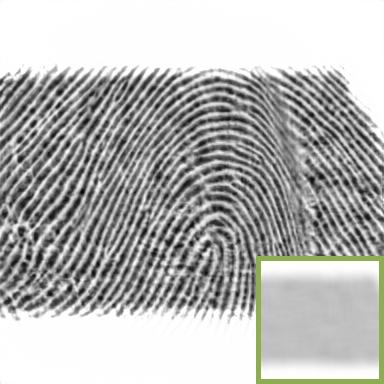}}\hfill &
    \fbox{\includegraphics[width=0.19\linewidth]{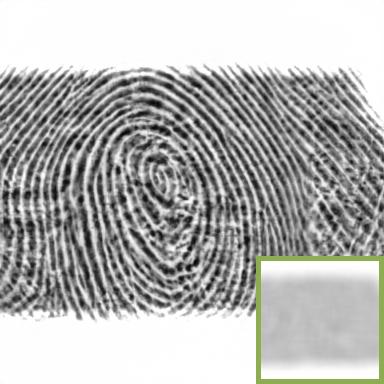}}\hfill &
    \fbox{\includegraphics[width=0.19\linewidth]{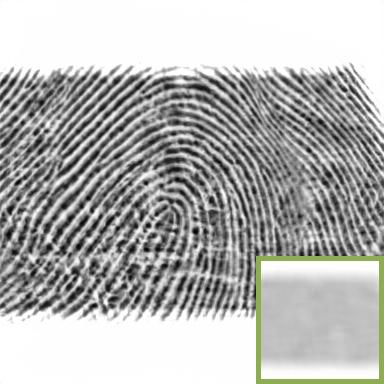}}\hfill &
    \fbox{\includegraphics[width=0.19\linewidth]{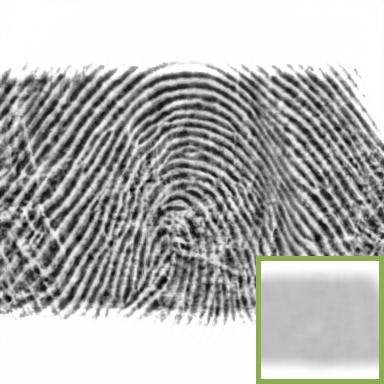}}\hfill &
    \fbox{\includegraphics[width=0.19\linewidth]{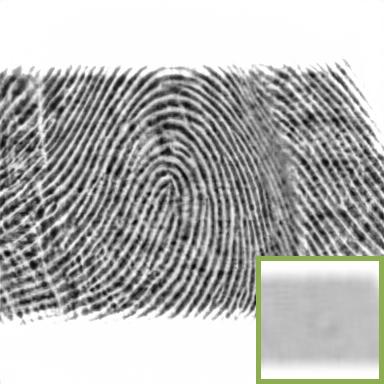}}\hfill \\
    \raisebox{0.7\totalheight}{\rotatebox[origin=c]{90}{\tiny Appearance 3}} &
    \fbox{\includegraphics[width=0.19\linewidth]{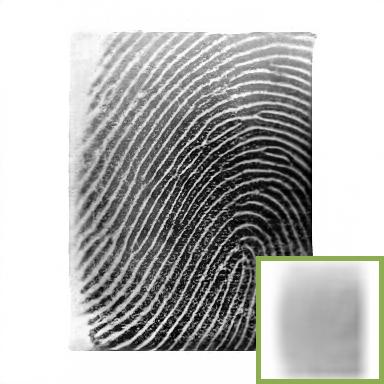}}\hfill &
    \fbox{\includegraphics[width=0.19\linewidth]{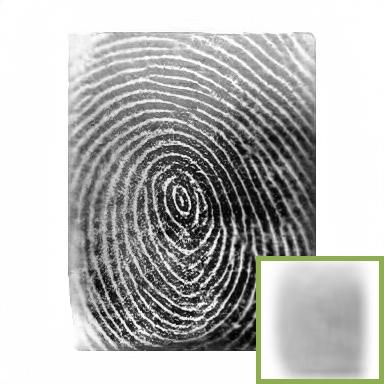}}\hfill &
    \fbox{\includegraphics[width=0.19\linewidth]{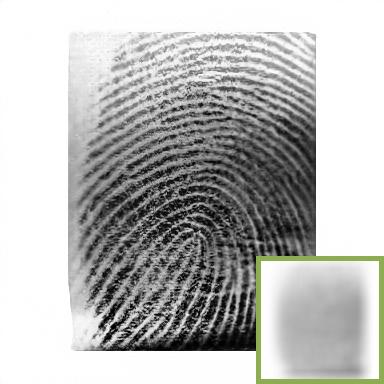}}\hfill &
    \fbox{\includegraphics[width=0.19\linewidth]{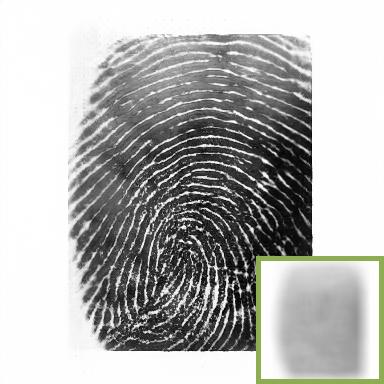}}\hfill &
    \fbox{\includegraphics[width=0.19\linewidth]{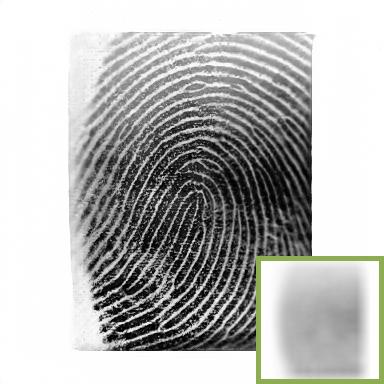}}\hfill \\

    \raisebox{0.7\totalheight}{\rotatebox[origin=c]{90}{\tiny Appearance 4}} &
    \fbox{\includegraphics[width=0.19\linewidth]{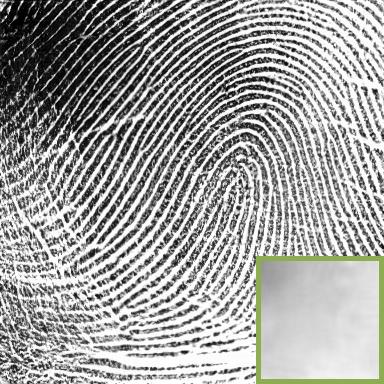}}\hfill &
    \fbox{\includegraphics[width=0.19\linewidth]{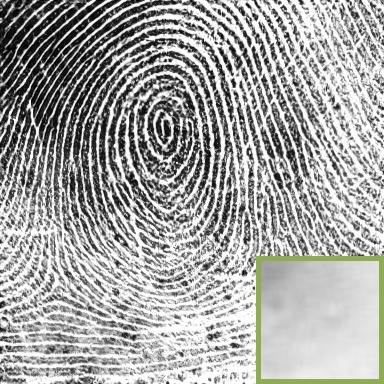}}\hfill &
    \fbox{\includegraphics[width=0.19\linewidth]{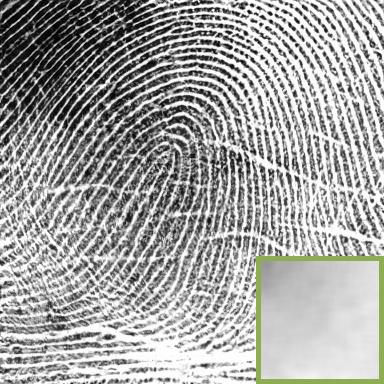}}\hfill &
    \fbox{\includegraphics[width=0.19\linewidth]{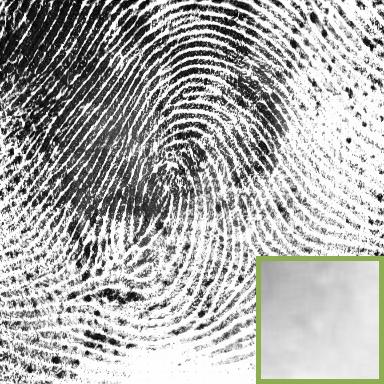}}\hfill &
    \fbox{\includegraphics[width=0.19\linewidth]{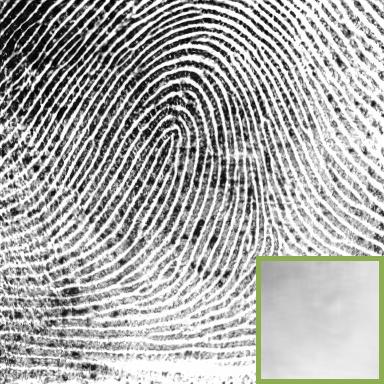}}\hfill \\
\end{tabular}
\caption{$w_{app}=20$.}
\label{fig:qualitative_results_20}
\end{subfigure}

\caption{
\textbf{Generation results of FPGAN-Control trained using different $w_{app}$.}
For a specific FPGAN-Control model, each column represents images generated with the same ID latent vector input and each row represents images generated with the same appearance latent vector input.
For visualization of the appearance loss, the small images in green borders show the blurred representation of the fingerprint image used by the loss.
}
\vspace{-1.0em}
\label{fig:qualitative_results}
\end{figure*}

%% file: experiments.tex
\section{Experiments}
\label{sec:experiments}
We first present quantitative and qualitative evaluations of FPGAN-Control, demonstrating its high quality generation, identity preservation and appearance control capabilities.
Second, we focus on the ability to train fingerprint recognition models on purely synthetic datasets generated by FPGAN-Control, which is our main goal.
We show that, given the multitude of identities and appearance variations that FPGAN-Control can generate, we are able to produce recognition models with higher accuracy even compared to models that were trained with real data.

In our experiments, we used the publicly available NIST N2N dataset (NIST SD 302~\cite{n2n}).
The N2N dataset contains 2,000 unique fingers (of 200 people) with 8 to 15 different impressions per finger.
Each finger represents a unique identity.
We divided the dataset to 1,600 identities (160 people) for training and 400 (40 people) for testing.
The N2N dataset is challenging for the task of training recognition models due to its diverse range of fingerprint images, captured using both traditional and newly developed methods.

We experimented with the following backbones: ResNet18, ResNet34, ResNet50, ResNet101~\cite{He2016DeepRL}, MobileNetV2-050, MobileNetV2-100~\cite{SandlerHZZC18} and EfficientNetV2-s~\cite{TanL21}.
As discussed in the previous section, we removed the first sub-sampling operation from all backbones to enhance training stability and improve accuracy. 
All recognition models were trained using the CosFace loss~\cite{Wang2018CosFaceLM}.
During training, we applied random affine transformations. 

\subsection{Synthetic fingerprint generation}
\label{sec:synthetic_fingerprint_generation}
In order to provide a recognition model for FPGAN-Control's identity loss, we first had trained a ResNet18-based model on the 1,600 identities of the training set. 
We used the same ResNet18 model for training all of our FPGAN-Control models.
Note that all our GANs were trained using only the 1,600 identities of the training set, the test set was never used in any form of training.
In the following sections, we will provide qualitative and quantitative results that demonstrate the capabilities of FPGAN-Conrol.

\subsubsection{Qualitative results}
Figure~\ref{fig:qualitative_results} shows qualitative results of four FPGAN-Control versions, each trained with a different appearance loss weight, $w_{app}$ ($0, 1, 5, 20$, where $0$ means no appearance loss).
Figure~\ref{fig:qualitative_results_0} demonstrates that without the appearance loss, the appearance variation between images of the same ID is small and mostly manifests changes in ridge pattern brightness.
Additionally, when $w_{app} = 0$, the generation exhibits no control over the appearance, having images generated with the same appearance latent look completely different.
Figures~\ref{fig:qualitative_results_1},~\ref{fig:qualitative_results_5},~\ref{fig:qualitative_results_20} demonstrate the gradual appearance control improvement when the $w_{app}$ is increased: from having small similarities when $w_{app}=1$ (\eg, some correlation between images in the second row of Figure~\ref{fig:qualitative_results_1}) to having almost identical appearances when $w_{app}=20$ (\eg, note the similar white triangle on the right corner of each image in the second row of Figure~\ref{fig:qualitative_results_20}).
Furthermore, Figure~\ref{fig:qualitative_results} shows that setting higher $w_{app}$ values increases the appearance variability for the same identity.
As an example, for $w_{app}=1$, ID3 has a similar appearance for all three appearance latent, while for every ID of $w_{app}=5$ and $w_{app}=20$ the appearances changes drastically for different appearance latent vectors.

\input{tables/table_id_app_same_id}
\input{tables/table_same_app}
\subsubsection{Intra-class distribution of generated images}
Next, we demonstrate the impact of $w_{app}$ on the intra-class variability of fingerprints generated by FPGAN-Control.
To measure the intra-class distribution,
for each FPGAN-Control version, we randomly generated 1,000 synthetic identities.
Each identity consists of two images generated using different appearance latents.
For each pair of images, we measured two distances: cosine distance between the embedding vectors of the two images (using a ResNet18 recognition model) and their appearance distance computed by equation ~\ref{eq:distance}.
Table~\ref{tab:id_app_same_id} summarizes the results and verifies the trends observed in the qualitative evaluation.
As the $w_{app}$ increases the appearance distance calculated between images of the same identity grows, implying that the variance in appearance of same identity fingerprints is increasing.



\subsubsection{Appearance control}
We quantify the ability to control the appearance of FPGAN-Control.
For each FPGAN-Control version, we randomly sampled 1,000 pairs of images generated with shared appearance latent and different identity latent. 
We computed the appearance distances, $d_{app}$, between all the pairs and report the results in Table~\ref{tab:same_app}.
As expected, FPGAN-Control models trained with larger $w_{app}$ exhibit better control over appearance.
These results support the qualitative results presented in Figure~\ref{fig:qualitative_results}, demonstrating the high level of appearance control achieved by FPGAN-Conrtol, where two images with different identity have nearly identical appearances.




\input{figures/mnt_corr}

\input{figures/minutiae_score_distributions}
\input{tables/table_50K_synth}
\subsubsection{Minutiae-points statistic}
Since the early days when fingerprints first began to be studied, minutiae-points (Figure~\ref{fig:mnt_corr}) have been a primary feature used to distinguish one fingerprint from another~\cite{galton}.
It is only in recent years that deep networks have been used to extract discriminative embeddings from fingerprints~\cite{deepprint}.
There are several benefits of using deep networks instead of minutiae matching approaches: (1) they allow faster matching\footnote{Minutiae matching approaches require expensive graph matching techniques.}~\cite{deepprint}, (2) enable matching in the encrypted domain~\cite{HERS2022} and (3) can still perform successful matches when the fingerprint quality is very low~\cite{deepprint}.
To further show our ability to control the identity of fingerprints (as defined by minutiae points) in the presence of various impression styles (appearances), we computed the minutiae similarity score distributions of our synthetic fingerprints. 
We used the open source minutiae matcher from~\cite{cao2019end}.
Figure~\ref{fig:minutiae_distributions} shows that as our appearance loss weight $w_{app}$ increases from $w_{app}=0$ to $w_{app}=20$, the minutiae similarity score distributions shift towards the distribution of minutiae scores computed from real fingerprints.
This lends additional strong evidence to our ability to maintain the identity of fingerprints as we modulate through different appearances.

\input{figures/res18_5k_50k}
\subsection{Training with synthetic data}   
In this section we report the accuracy results obtained by recognition models trained using synthetic images that were generated by FPGAN-Control.
We evaluated the performance of the trained recognition models on the test subset of the N2N dataset (real data).
For training, we assumed that the real training dataset is no longer available and only synthetic identities were involved in the training process.
To generate a synthetic identity, we have randomly sampled one common ID latent vector, $\mathbf{z}^{id}$, and 11 different appearance latent vectors, $\{ \mathbf{z}^{app}_i \}_{i=1}^{11}$.
We then concatenated the ID latent vector to each one of the appearance latent vectors, \ie $[\mathbf{z}^{id}, \mathbf{z}^{app}_i]$.
Finally, we provide the concatenated vectors as input to FPGAN-Control, generating 11 fingerprint images of the same identity, each having a different appearance.
A synthetic dataset is constructed by generating multiple such synthetic identities. 

In Table~\ref{tab:50k_synth}, we present the accuracy results (measured by TAR@FAR=0.1\%) of various recognition models trained on different synthetic datasets using different network architectures.
Specifically, we compare the results obtained by our FPGAN-Control to two baseline synthetic datasets.
In the first, a regular StyleGAN2~\cite{karras2020analyzing} model was used to generate a multitude of synthetic fingerprint images.
Then, each individual image was duplicated 11 times to define a unique identity.
The second baseline is the publicly available dataset created by the PrintsGAN approach~\cite{engelsma2022printsgan}.
Multiple datasets were generated by the FPGAN-Control approach, each of which was named after the weight of its appearance loss. 
For example, FPGC-5 dataset was generated by setting $w_{app}=5$ during the training of FPGAN-Control.
We also combined images generated by multiple FPGAN-Control models to increase the diversity of the synthetic dataset. 

From Table~\ref{tab:50k_synth}, we first observe a significant improvement of the proposed FPGAN-Control approach compared to the baseline datasets generated by the StyleGAN2 model and the PrintsGAN approach. 
Specifically, models trained using StyleGAN2's data struggled to converge, and training with PrintsGAN's data yielded poor recognition results. 
The significant leap in recognition accuracy obtained by FPGAN-Control demonstrates the effectiveness of the proposed approach in generating reliable fingerprint images that are useful for the task of training recognition models.

Secondly, in most cases, setting higher weight for the appearance loss of the FPGAN-Control model results in higher accuracy. 
For example, training with FPGC-20 is superior for the ResNet backbone family. 
In some cases, training a model using FPGC-20 can even achieve similar results compared to training with the original real data (e.g. training with 50K synthetic identities using ResNet-34 backbone). 
This shows the importance of the disentanglement between identity and appearance information that enables an increase in the variability of the fingerprint images generated by the FPGAN-control model.

In Figure~\ref{fig:res18_5k_50K} we present the TAR@FAR=0.1\% and TAR@FAR=0.01\% while gradually increasing the number of synthetic identities used during the training of the recognition model from 5K to 80K.
We show that, as we increase the number of identities, the performance increases until it either plateaus or begins to deteriorate.
We also report the results obtained by recognition models trained using PrintsGAN.
The recognition accuracies obtained by models trained on data generated with PrintsGAN are inferior compared to any of the recognition models trained on data generated with FPGAN-Control.
This might be partially due to the lack of sufficient appearance variation generated by PrintsGAN. 
Note that, for PrintsGAN, we evaluate the models using up to 30K synthetic identities, which constitute all the publicly available data released by the authors.

By combining multiple datasets generated by different FPGAN-Control models we obtain significant improvement of the recognition accuracy.
For example, when using 80K synthetic identities, ResNet18 is able to achieve TAR@FAR=0.1\% = 92.43\%, an improvement of 1.61\% compared to model trained by the real data.
Incorporating images generated by various FPGAN-Control models further increases the variability of the images used for training, which in turn leads to better accuracy results.

%% file: tables/table_id_app_same_id.tex
\begin{table}[t]
    \centering
    \small
    \begin{tabular}{l | c c}
    \toprule
    
    $w_{app}$ \quad\quad\quad\quad\quad\quad\quad\quad & \quad ID distance \quad & \quad App. distance \quad \\
    \midrule
    $0$    & $0.063$\tiny{$\pm0.06$}     & $0.026$\tiny{$\pm0.04$}  \\
    $0.25$ & $0.074$\tiny{$\pm0.07$}     & $0.027$\tiny{$\pm0.04$} \\
    $0.5$  & $0.076$\tiny{$\pm0.07$}     & $0.033$\tiny{$\pm0.04$} \\
    $1$    & $0.122$\tiny{$\pm0.08$}     & $0.043$\tiny{$\pm0.04$} \\
    $5$    & $0.229$\tiny{$\pm0.11$}     & $0.051$\tiny{$\pm0.05$} \\
    $20$   & $0.283$\tiny{$\pm0.11$}     & $0.057$\tiny{$\pm0.05$} \\
    \midrule
    Real data & $0.375$\tiny{$\pm0.19$} & $0.058$\tiny{$\pm0.05$}  \\
    \bottomrule
    \end{tabular}
    \caption{\textbf{Intra class statistic.}
    ID distance and App distance correspond to the average recognition distance and the mean appearance distance between two images of the same ID, respectively.}
    \label{tab:id_app_same_id}
\end{table}

%% file: tables/table_same_app.tex
\begin{table}[t]
    \centering
    \small
    \begin{tabular}{l c c c c c c}
    \toprule
    $w_{app}$   & $0$    & $0.25$ & $0.5$ & $1$ & $5$ & $20$ \\
    \midrule
    \raisebox{-3pt}{Dist$\downarrow$} & $\underset{\pm0.04}{0.053}$ & $\underset{\pm0.04}{0.044}$ & 
    $\underset{\pm0.03}{0.038}$ & $\underset{\pm0.02}{0.025}$ & $\underset{\pm0.01}{0.009}$ & 
    $\underset{\pm0.00}{0.002}$ \\
    \bottomrule
    \end{tabular}
    \caption{\textbf{Appearance control precision vs. appearance loss weight, $w_{app}$}.
    For each FPGAN-Control model, we measured the average appearance distance between pairs of images sharing the same appearance latent, but having a diffrent ID latent.} 
    \label{tab:same_app}
\end{table}




%% file: figures/mnt_corr.tex
\begin{figure}[]
\centering
\renewcommand{\arraystretch}{1}

\includegraphics[width=0.9\linewidth]{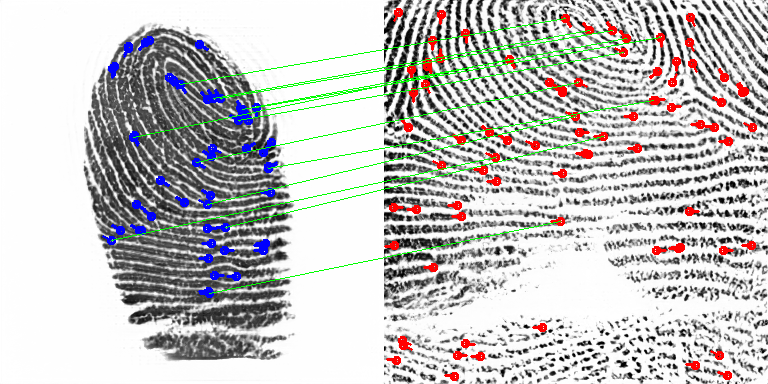}
\includegraphics[width=0.9\linewidth]{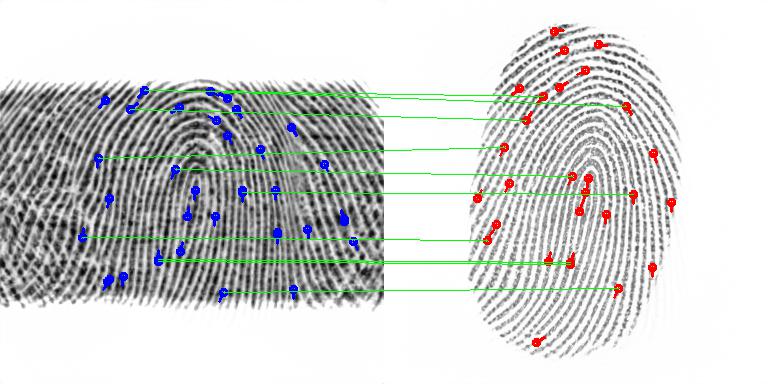}
\vspace{-0.3cm}
\caption{
\textbf{Minutiae matching quality.}
Examples of minutiae matching on genuine pairs (same identity per row) of our synthetic fingerprints when $w_{app}=20$. The locations of minutiae points are annotated by circles and the orientations are indicated by the tails appended to each circle. The minutiae matcher then aims to find as many corresponding minutiae points as possible across the pairs. We note that even after placing heavy weight on our appearance loss, the minutiae defined identities are maintained across the image pairs as indicated by a large number of correspondences established.
}
\vspace{-0.6em}
\label{fig:mnt_corr}
\end{figure}

%% file: figures/minutiae_score_distributions.tex
\begin{figure}
\centering
\includegraphics[width=0.95\linewidth]{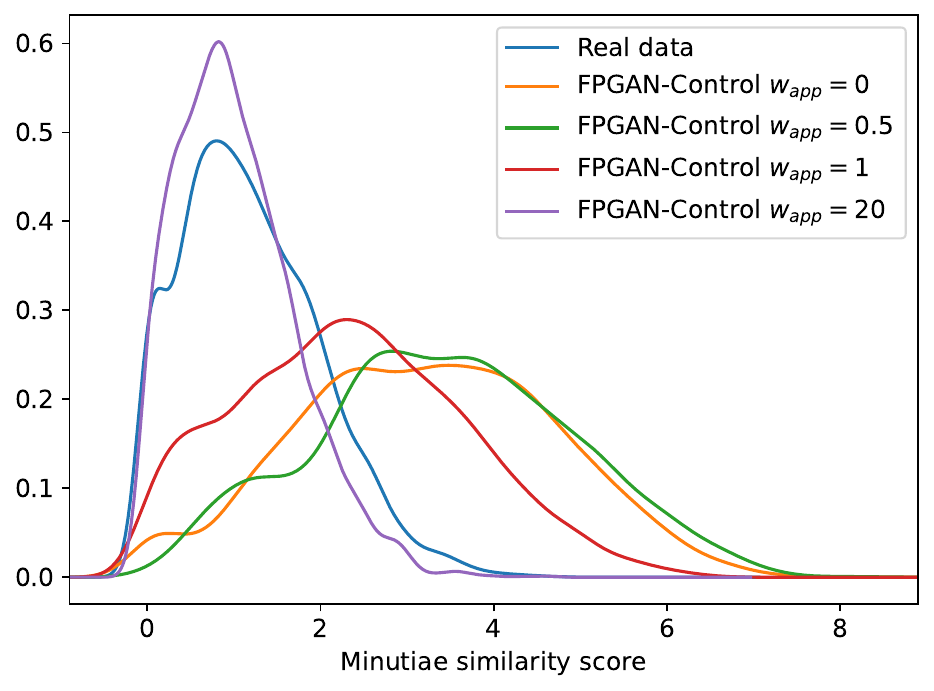}
\caption{\textbf{Minutiae similarity score distributions.}}
\vspace{-1.5em}
\label{fig:minutiae_distributions}
\end{figure}

%% file: tables/table_50K_synth.tex
\begin{table*}[t]
    \centering
    
    \newcolumntype{C}[1]{>{\centering\arraybackslash}p{#1}}
    \begin{tabular}{l | C{0.08\textwidth} C{0.08\textwidth} C{0.08\textwidth} C{0.08\textwidth} C{0.08\textwidth} C{0.08\textwidth} C{0.08\textwidth}}
        \toprule
        \textbf{Training dataset}   & \textbf{Res18}          & \textbf{Res34}          & \textbf{Res50}         & \textbf{Res101}        & \textbf{Mob-050} & \textbf{Mob-100} & \textbf{Eff-s} \\ 
        \midrule
        Real data            & $90.54$        & $89.72$        & $92.00$       & $93.69$       & $93.74$ & $94.22$ & $93.69$ \\
        \midrule
        StyleGAN2                & $23.54$        & $6.085$        & $6.47$        & $8.87$        & $30.70$ & $32.37$ & $5.59$     \\
        PrintsGAN*                & $69.61$        & $63.09$        & $71.65$       & $72.26$       & $73.99$ & $79.09$ & $67.83$    \\ 
        FPGC-0     & $87.53$        & $87.25$        & $87.01$       & $87.60$       & $88.72$ & $90.13$ & $83.87$    \\
        FPGC-0.25 &  $86.13$        & $85.83$        & $87.74$          &  $89.01$       & $88.13$ & $90.75$ & $80.40$     \\
        FPGC-0.5  &  $88.00$       & $87.24$        & $86.71$          &  $87.26$       & $88.20$ & $90.50$ & $80.72$     \\
        FPGC-1   &  $89.42$       & $87.70$        & $88.60$          &  $89.51$       & $\textbf{90.63}$ & $\textbf{91.14}$ &  $85.20$ \\
        FPGC-5   &  $87.48$       & $86.97$        & $87.81$          &  $90.15$       & $88.72$ & $89.68$ &  $82.82$    \\
        FPGC-20  & $\textbf{89.57}$ & $\underline{\textbf{89.72}}$ & $\textbf{90.19}$ &  $\textbf{91.08}$      & $89.99$ & $90.88$ &  $\textbf{88.55}$ \\
        \midrule
        FPGC-0.25 + FPGC-20 &  $\underline{91.60}$          & $\underline{91.59}$             & $91.99$          & $90.70$        & $92.53$ & $93.22$ & $88.96
$     \\
        FPGC-0.5 + FPGC-20   &  $\underline{\textbf{92.15}}$ & $\underline{\textbf{92.24}}$    & $\underline{92.47}$          & $91.36$        & $\textbf{92.62}$ & $92.78$ &  $\textbf{90.60}$    \\
        FPGC-1 + FPGC-20     &  $\underline{91.86}$          & $\underline{91.88}$             & $\underline{\textbf{92.58}}$ & $87.33$        & $91.98$ & $\textbf{93.23}$ &  $89.88$    \\
        FPGC-5 + FPGC-20    &  $\underline{91.22}$          & $\underline{91.06}$             & $91.88$          & $\textbf{91.69}$        & $91.51$ & $92.21$ & $89.76$     \\

        \bottomrule
    \end{tabular}
    \caption{    
    \textbf{Recognition results for 50K synthetic identities.}
    TAR@FAR=0.1\% results obtained by recognition models with various backbones trained using different synthetic datasets for the case of 50K synthetic identities. 
    The datasets that were generated by FPGAN-control are denoted by FPGC-$w_{app}$ where $w_{app}$ corresponds to the weight of the appearance loss.
    We use an \underline{underline} to denote models that surpass or are equal to the performance of models trained on real data.
    }
    \label{tab:50k_synth}
\end{table*}

%% file: figures/res18_5k_50k.tex
\begin{figure*}[h]
\centering
\begin{subfigure}[t]{0.49\linewidth}
\includegraphics[width=0.9\linewidth]{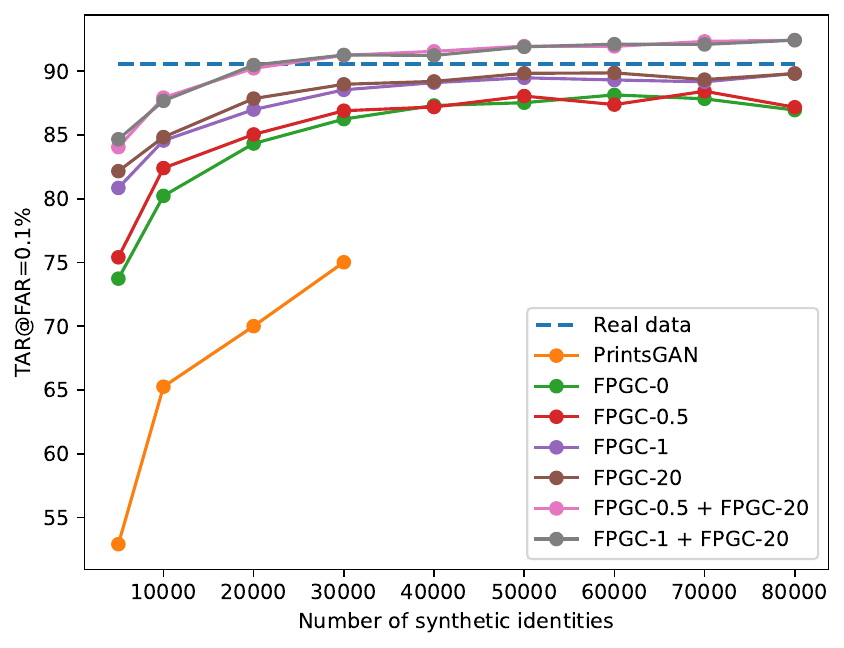}
\caption{ResNet18 TAR@FAR $=0.1\%$.}
\end{subfigure}
\begin{subfigure}[t]{0.49\linewidth}
\includegraphics[width=0.9\linewidth]{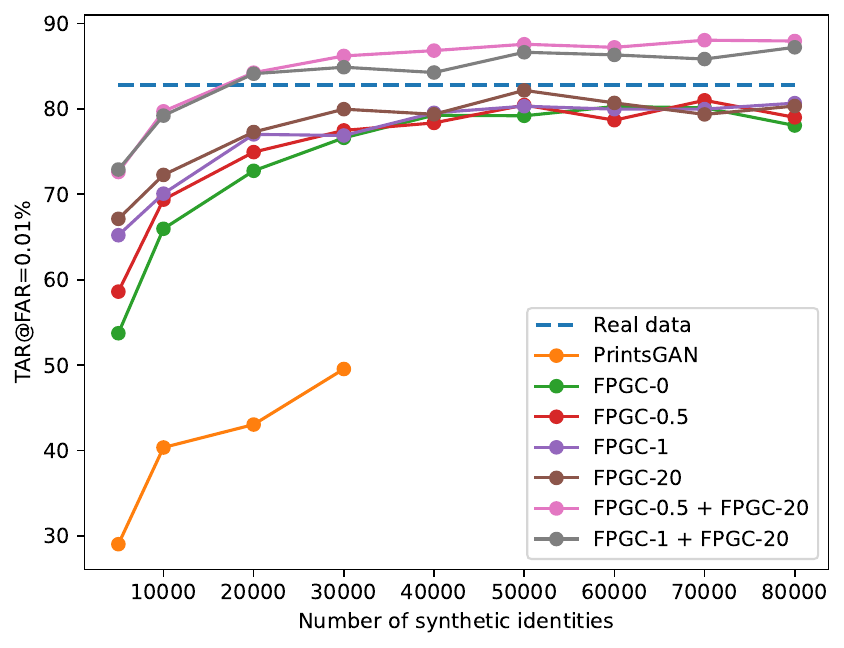}
\caption{ResNet18 TAR@FAR $=0.01\%$.}
\end{subfigure}

\caption{
\textbf{Accuracy vs. number of synthetic identities used during training.}
Real data corresponds to training the model with the entire 1,600 identities real dataset only, while the rest of the models were trained purely on synthetic identities. Note that PrintsGAN published only 35K identities, all of which were used in this evaluation. 
}
\vspace{-1.0em}
\label{fig:res18_5k_50K}
\end{figure*}

%% file: conclusions.tex
\section{Conclusions}
\label{sec:conclusions}
We presented FPGAN-Control, a novel framework for training fingerprint generation models which can synthesize multiple images of the same novel fingerprint identity while controlling its appearance.
We introduced a novel appearance loss for disentangling FPGAN-Control's latent space enabling control over generated fingerprint appearance while preserving their identity. 
Finally, the datasets generated by FPGAN-Control were used to train recognition models relying solely on synthetic identities and we showed that we are able to reach comparable and even higher accuracies than models trained using real data only.